\documentclass[3p,preprint,10pt]{elsarticle}
\usepackage{amsmath,amssymb,amsfonts}
\usepackage{bm}
\usepackage{graphicx}
\usepackage{textcomp}
\usepackage{array,booktabs}
\usepackage[caption=false, font={scriptsize,sf},subrefformat=parens]{subfig}
\usepackage[ruled,norelsize]{algorithm2e}
\SetAlCapFnt{\small}
\SetAlCapNameFnt{\small}
\SetAlFnt{\small}

\newcommand{\figref}[1]{\figurename~\ref{#1}}

\journal{Engineering Applications of Artificial Intelligence}

\begin{document}
\begin{frontmatter}

\title{A Comprehensive End-to-End Computer Vision Framework for Restoration and Recognition of Low-Quality Engineering Drawings}

\author[hust]{Lvyang~Yang}
\ead{hkyly_217@hust.edu.cn}

\author[sgcc]{Jiankang~Zhang}
\ead{zhangjk@nw.sgcc.com.cn}

\author[sgcc]{Huaiqiang~Li}
\ead{lihq@nw.sgcc.com.cn}

\author[sgcc]{Longfei~Ren}
\ead{ren.longfei@qq.com}

\author[hust]{Chen~Yang}
\ead{yangchen1003@hust.edu.cn}

\author[hust]{Jingyu~Wang\corref{corr}}
\ead{jywang@hust.edu.cn}

\author[hust]{Dongyuan~Shi}
\ead{dongyuanshi@hust.edu.cn}

\cortext[corr]{Corresponding author}
\affiliation[hust]{
    organization={State Key Laboratory of Advanced Electromagnetic Technology, Huazhong University of Science and Technology},
    city={Wuhan},
    postcode={430074}, 
    state={Hubei},
    country={China}}
\affiliation[sgcc]{
    organization={Northwest Branch of State Grid Corporation of China}, 
    city={Xi'an},
    postcode={710048}, 
    state={Shaanxi}, 
    country={China}}

\begin{abstract}
The digitization of engineering drawings is crucial for efficient reuse, distribution, and archiving. Existing computer vision approaches for digitizing engineering drawings typically assume the input drawings have high quality. However, in reality, engineering drawings are often blurred and distorted due to improper scanning, storage, and transmission, which may jeopardize the effectiveness of existing approaches. This paper focuses on restoring and recognizing low-quality engineering drawings, where an end-to-end framework is proposed to improve the quality of the drawings and identify the graphical symbols on them. The framework uses K-means clustering to classify different engineering drawing patches into simple and complex texture patches based on their gray level co-occurrence matrix statistics. Computer vision operations and a modified Enhanced Super-Resolution Generative Adversarial Network (ESRGAN) model are then used to improve the quality of the two types of patches, respectively. A modified Faster Region-based Convolutional Neural Network (Faster R-CNN) model is used to recognize the quality-enhanced graphical symbols. Additionally, a multi-stage task-driven collaborative learning strategy is proposed to train the modified ESRGAN and Faster R-CNN models to improve the resolution of engineering drawings in the direction that facilitates graphical symbol recognition, rather than human visual perception. A synthetic data generation method is also proposed to construct quality-degraded samples for training the framework. Experiments on real-world electrical diagrams show that the proposed framework achieves an accuracy of 98.98\% and a recall of 99.33\%, demonstrating its superiority over previous approaches. Moreover, the framework is integrated into a widely-used power system software application to showcase its practicality. The reference codes and data can be found at https://github.com/Lattle-y/AI-recognition-for-lq-ed.git.
\end{abstract}


\end{frontmatter}

\section{Introduction}
\label{sec:introduction}

Engineering drawings (EDs) are crucial design documents utilized across various engineering disciplines, including electrical, mechanical, and civil engineering, to name a few. They use graphical elements, such as geometric shapes, lines, and text annotations, to convey the technical details of a particular engineering object. In the pre-digital era, most EDs were in paper form, resulting in challenges for efficient retrieval and reuse, as well as preservation issues such as aging and fading. With the increasing application of automation technology in engineering projects, there is a growing need to digitize EDs. Initially, ED digitization involved scanning paper-based EDs into images, which addressed preservation concerns but did not enhance the efficiency of engineering practices. The scanned EDs, being raster images, are not directly parseable by most computer-aided engineering software. Recently, significant breakthroughs in artificial intelligence (AI) techniques for computer vision (CV) have emerged, enabling the automatic transformation of raster ED images into parseable digital description files. For instance, Song \textit{et al.} propose a deep learning-based automatic electrical drawings recognition system in \cite{song2021edrs}, which uses a Faster Region-based Convolutional Neural Network (Faster R-CNN) model to detect electrical symbols. Similarly, Yang \textit{et al.} propose an intelligent CV framework in \cite{yang2023intelligent} to digitize power substation one-line diagrams, which uses a You Only Look Once (YOLO) model to detect and classify electrical device icons and a combination of a Character Region Awareness For Text detection (CRAFT) model and a Convolutional Recurrent Neural Network (CRNN) to locate and identify text labels. Additionally, Nurminen \textit{et al.} use a YOLO model in \cite{nurminen2020object} to detect high-level objects such as pumps and valves in Piping and Instrumentation Diagrams (P\&IDs). Elyan \textit{et al.} propose an advanced bounding-box detection method for ED symbol recognition in \cite{elyan2020deep}, integrating a Generative Adversarial Network (GAN) model to handle the class imbalance problem of different types of symbols. These AI-based ED digitization applications have exhibited reasonable recognition accuracy, significantly enhancing the efficiency of ED retrieval, maintenance, and reuse when compared to simple scanning methods. However, a drawback of these applications is their dependence on high-quality EDs for training learning models, making them less adept at handling low-quality EDs. When the input ED images suffer from issues such as low resolution, blur, or noise, the recognition performance is notably compromised. 

The prevalence of low-quality EDs in practical scenarios exacerbates this drawback. On the one hand, the performance limitations of early scanners led to the generation of low-resolution ED images in the past. Furthermore, during that period, constraints in disk space and communication bandwidth often resulted in additional compression during storage and transmission of ED images. On the other hand, paper-based ED documents are susceptible to damage over time, and even with contemporary high-definition scanners, preservation issues such as ink fading, dirt, and dust can introduce blurriness and noise, which detrimentally impact the identification of graphical symbols. Therefore, enhancing the recognition performance of low-quality EDs is a pivotal technical challenge that needs to be addressed to broaden the applicability of AI-based ED digitization approaches.

To address the challenge, this paper focuses on the restoration and recognition of low-quality EDs. It proposes an end-to-end framework consisting of an image restoration module and a graphical symbol recognition module, with a series of deep learning CV techniques integrated to improve the digitization performance of complex EDs with severe quality degradation issues. The primary contributions of this paper include:
\begin{enumerate}
  \item To enhance the efficiency of the image restoration module, this paper employs K-means clustering to categorize diverse ED patches into Simple Texture Patches (STPs) and Complex Texture Patches (CTPs) according to the Gray Level Co-occurrence Matrix (GLCM) statistics. For the substantial number of STPs, simple CV heuristics are applied to improve image quality at a reasonable computational expense. Conversely, for the limited number of CTPs, a modified Enhanced Super-Resolution Generative Adversarial Network (ESRGAN) model is utilized to ensure high-quality restoration performance.
  \item To improve the accuracy of ED graphical symbol recognition, this paper refines the widely-used Faster R-CNN model by incorporating Swin Transformers as the backbone and integrating Bi-directional Feature Pyramid Networks (BiFPN) to enhance its multi-scale feature extraction capability. Additionally, a multi-stage task-driven collaborative learning strategy is proposed to train the modified ESRGAN and Faster R-CNN models as a whole, maximizing the contribution of super-resolution to symbol recognition.
  \item Following \cite{song2021edrs} and \cite{yang2023intelligent}, electrical engineering is chosen as a representative application field to evaluate the proposed framework. A comprehensive assessment is conducted using real-world electrical diagrams supplied by our power utilities partners. Additionally, the framework is integrated into RelayCAC, a widely-used commercial power system protective relay setting coordination software in China, to showcase its practical application capability.
\end{enumerate}

The remainder of the paper is organized as follows. Section \ref{sec:related_work} provides a brief review of the related work. Section \ref{sec:preliminary_technology} introduces the CV techniques utilized in the proposed ED digitization framework. Section \ref{sec:low_quality_ed_recognition_framework}
 presents the overall architecture and describes the constituent modules in detail. Section \ref{sec:experiments_and_discussions} analyzes the evaluation results. Finally, Section \ref{sec:conclusion} concludes the paper.

\section{Related Work}
\label{sec:related_work}

This section briefly reviews the relevant literature, covering research on AI-based ED symbol detection and recognition, restoration techniques for low-quality images, and recent advancements in image recognition models designed for quality-degraded images.

\subsection{ED Symbol Detection and Recognition}
\label{sub:ed_symbol_detection_and_recognition}

ED symbol detection and recognition have long been prominent research areas within CV and pattern recognition. Early traditional methods typically combine image processing and feature extraction algorithms, such as edge detection and texture extraction algorithms, with classic machine learning models to detect ED symbols \cite{escalera2010circular,yu1997system,llados2001symbol}. These methods are efficient and can achieve decent detection performance on simple EDs, and hence have long-lasting influence. Even in recent years, Kang \textit{et al.} still employ template matching, an image processing technique, to detect symbols on P\&IDs \cite{kang2019digitization}. Nonetheless, these methods have obvious drawbacks. They are often empirically designed and lack generalizability to different symbol styles, leading to limited applicability in real-world scenarios. 

The advent of deep learning, particularly Convolutional Neural Networks (CNNs), has reshaped the research on symbol detection and recognition, rendering it more flexible and accurate. In \cite{moreno2019new}, Moreno-García \textit{et al.} propose a comprehensive framework for digitizing complex P\&IDs. They also provide a thorough review of relevant machine learning models and discuss how emerging trends in deep learning could be applied to solve this problem. In \cite{pizarro2022automatic}, Pizarro \textit{et al.} present a survey of rule-based and learning-based approaches for recognizing architectural floor plan diagrams. Additionally, Sarkar \textit{et al.} introduce an ED symbol recognition method using multiple deep neural networks in \cite{sarkar2022automatic}. This method autonomously learns the shapes of symbols from the legend table without requiring specific training with predefined templates, demonstrating its generality and domain independence. Deep learning techniques have also found application in the detection and recognition of handwritten symbols \cite{paul2022ensemble} and flowchart symbols \cite{schafer2022sketch2process}, significantly enhancing their accuracy and robustness.

\subsection{Restoration of Low-Quality Images}
\label{sub:restoration_of_low_quality_images}

The aforementioned research explicitly requires or implicitly assumes that the input EDs have high quality. When the input EDs exhibit degraded quality, the accuracy and robustness of these methods can be significantly compromised. This is because reduced image resolution can blur symbol boundaries, while dirt and noise can obscure details. Additionally, fading and distortion can further impact feature integrity. To address these adverse effects of image quality degradation on graphical symbol recognition, an intuitive idea is to restore high-quality images from low-quality images before feeding them into the subsequent recognition application.

The restoration of low-quality images typically involves three aspects: image denoising, deblurring, and super-resolution (SR). Babu and Tian \cite{babu2021review,tian2020deep} review the literature applying deep neural networks to image denoising and comfirm their superior performance over traditional methods. Zhang \textit{et al.} \cite{zhang2022deep} discuss deep learning-based image deblurring approaches to recover sharp images from blurred inputs. Datta \textit{et al.} \cite{datta2022single} propose a single-image SR approach based on sparse representation, demonstrating favorable results on large-scale electrical, mechanical, and civil building design images. Ledig \textit{et al.} \cite{ledig2017photo} introduce the first GAN-based method for SR (SRGAN), displaying strong performance on natural images for 4\texttimes upscaling factors. Zamir \textit{et al.} \cite{zamir2021multi} propose a multi-stage image restoration architecture, including deblurring, denoising, and decontamination, which achieves performance gains across various real-world tasks. However, it is important to note that while these methods enhance image quality to facilitate human visual perception, they often inadequately consider the relationship between low-quality image restoration and downstream tasks such as symbol detection and recognition.

\subsection{Image Recognition Considering Quality Degradation}
\label{sub:image_recognition_considering_quality_degradation}

Several studies adopt a holistic approach to address both quality enhancement and image recognition. In \cite{pei2020consistency}, Pei \textit{et al.} propose a consistency-guided network that realizes enhanced classification performance on quality-degraded images by aligning the semantic information regions in clear and degraded images. Endo \textit{et al.} \cite{endo2020cnn} introduce a CNN-based ensemble learning framework that automatically infers ensemble weights based on the estimated degradation levels and the features of restored images, achieving robust classification across various degradation levels. In \cite{pang2019jcs}, Pang \textit{et al.} present a unified network integrating the classification and super-resolution (SR) tasks to detect small-scale pedestrians. Learning from large-scale pedestrians, the SR sub-network can recover detailed information meaningful for classification. In \cite{haris2021task}, Haris \textit{et al.} propose a deep neural network where the SR sub-network explicitly incorporates a detection loss in its training objective, enabling to use any differentiable detector to train the SR sub-network. However, the loss function and the training strategy are not presented in detail. Bai \textit{et al.} \cite{bai2018sod} introduce a multi-task GAN for small object detection, which uses an SR network as the generator and a multi-task network as the discriminator. Classification and regression losses in the discriminator are back-propagated into the generator during training to enhance the generator's feature extraction capability for easier object detection. Moreover, GAN-based SR techniques are combined with object detection models in the realm of remote sensing for processing low-resolution photographs \cite{rabbi2020small,wang2022remote}.

While the aforementioned studies integrate image quality enhancement with downstream tasks like image recognition, their focus on natural images makes their direct application to complex EDs challenging. Unlike natural images, EDs usually exhibit less contextual information and are predominantly binarized, lacking depth of field information. In addition, graphical symbols on EDs often display high imbalance in type and size, posing difficulties for recognition models in equally addressing each variation. Given these distinctions, it is imperative to leverage insights from existing research and devise an end-to-end restoration and recognition framework specifically tailored for processing low-quality EDs.

\section{Preliminary Technology}
\label{sec:preliminary_technology}

This section provides an overview of the foundational techniques integral to the proposed framework, including GLCM for characterizing patch texture complexity, ESRGAN for image restoration, and Faster R-CNN for symbol detection.

\subsection{GLCM for Grayscale Texture Characterization}
\label{sub:glcm_for_grayscale_texture_characterization}

GLCM refers to the gray-level co-occurrence matrix, which is widely used to describe the spatial relationships of different pixels in a grayscale image. In simpler terms, it represents how often different combinations of pixels with specific gray levels and specific relative positions occur in an image. To compute a GLCM, the input grayscale image should be first quantized into $N$ discrete gray levels. Then, given an offset $(\Delta x, \Delta y)$, the $(i, j)$-th element of the corresponding GLCM $G_{\Delta x, \Delta y}$ can be calculated as
\begin{equation}
  G_{\Delta x, \Delta y}(i, j) = \sum_{x,y} \begin{cases}
    1 & \text{if $V(x,y)=i$ and $V(x + \Delta x, y + \Delta y)=j$} \\
    0 & \text{otherwise} 
    \label{eq:glcm}
  \end{cases}
\end{equation}
where $(x,y)$ and $(x+\Delta x, y+\Delta y)$ denote two pixels in the images, and the function $V(x,y)$ returns the gray level of the pixel $(x,y)$. The offset $(\Delta x, \Delta y)$ reflects the relative position of the two pixels, so choosing different offsets results in different GLCMs. 

GLCMs offer insights into the local variability of pixel values within an image. In regions with smooth textures, the grayscale values around a pixel typically exhibit a moderate range, while in areas with rough textures, this range is often more extensive. Consequently, the complexity of an image patch can be effectively characterized by computing statistical measures from GLCMs. Four representative statistical measures, namely dissimilarity $M_d$, homogeneity $M_h$, energy $M_e$, and entropy $M_p$, can be computed as follows:
\begin{equation}
    M_d = \sum_{i,j=0}^{N-1} |i-j|P_{ij}
    \label{eq:dissimilarity}
\end{equation}
\begin{equation}
    M_h = \sum_{i,j=0}^{N-1}\frac{P_{ij}}{1+(i-j)^2}
\end{equation}
\begin{equation}
    M_e = \sum_{i,j=0}^{N-1} P_{ij}^2
\end{equation}
\begin{equation}
    M_p = \sum_{i,j=0}^{N-1}-\ln(P_{ij})P_{ij}
    \label{eq:entropy}
\end{equation}
where $P_{ij}$ represents the $(i,j)$-th element of the normalized GLCM, which is obtained by dividing the raw GLCM by the sum of its element values. That is to say, $\sum_{i,j}^{N-1}P_{ij}=1$.

Among these statistical measures, dissimilarity quantifies the degree of grayscale differences between pixel pairs in the region of interest. If the image patch has a constant gray level, dissimilarity will equal zero. Homogeneity gauges the proximity of the observed GLCM distribution to that of a diagonal GLCM. A diagonal GLCM implies that the pixels $(x, y)$ and $(x+\Delta x, y+\Delta y)$ share the same gray level for any $(x, y)$, indicating a regularly repeated texture in the image patch. Energy calculates the sum of squared elements in a GLCM, reflecting the uniformity of the image patch. In contrast, entropy assesses the randomness of the pixel pair distribution in the image patch. Since these measures offer diverse perspectives on characterizing image textures, they are integrated into the proposed framework to serve as input features for a K-means clustering model to classify different ED patches into STPs and CTPs.

\subsection{ESRGAN for Degraded Image Restoration}
\label{sub:esrgan_for_degraded_image_restoration}

\begin{figure}[!htb]
  \centering
  \includegraphics{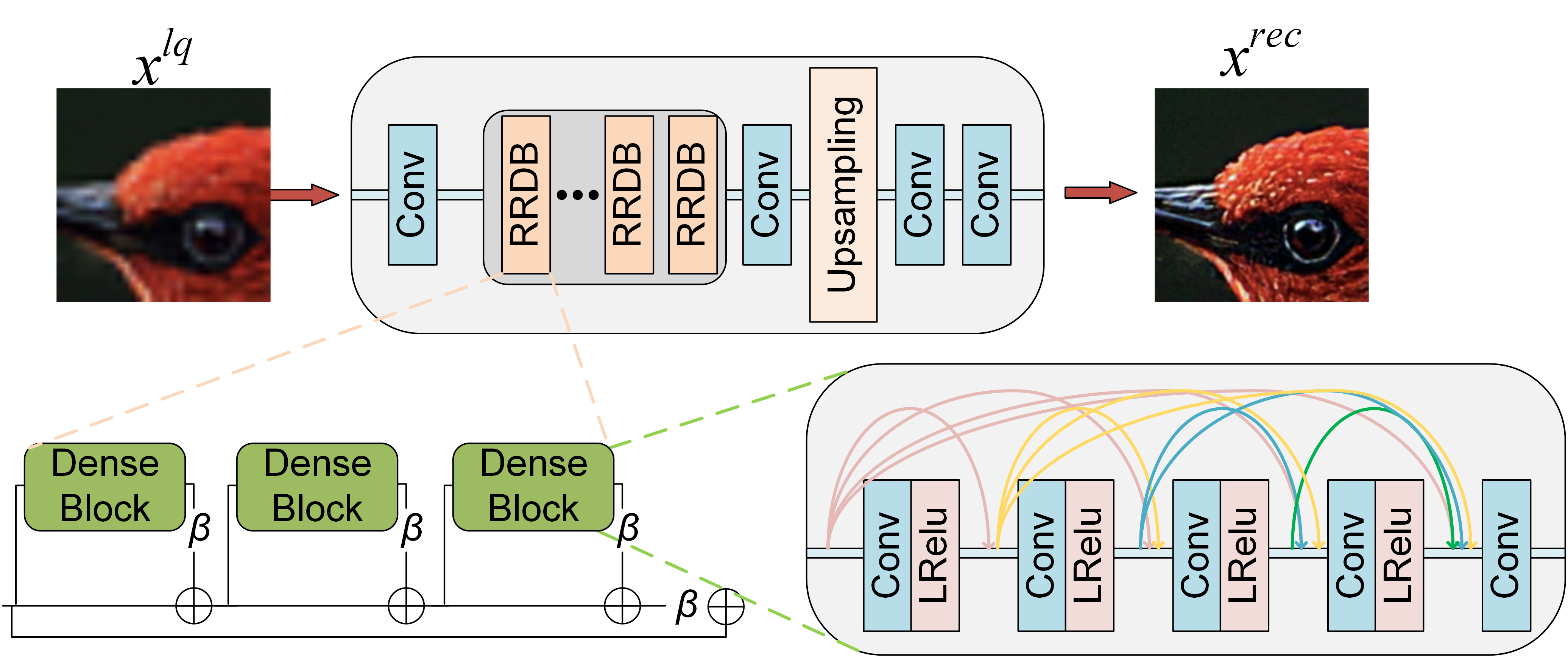}
  \caption{The network structure of the generator in ESRGAN.}
  \label{fig:esrgan}
\end{figure}

ESRGAN \cite{wang2018esrgan} is an advanced SR model that builds upon the SRGAN model \cite{ledig2017photo}. Although initially designed to enhance the visual quality of natural images, it is extended in the proposed framework to the recovery of low-quality EDs. ESRGAN comprises two networks: a generator and a discriminator. The generator takes a low-resolution image as input and produces the corresponding SR image. As shown in \figref{fig:esrgan}, it adopts the basic architecture of CNN and replaces the original residual block in SRGAN with a multi-level residual network known as Residual-in-Residual Dense Block (RRDB). In contrast to the residual block, RRDB avoids the use of batch normalization layers, as they may generate undesirable artifacts when the patterns of the training and test images differ significantly. The discriminator is a binary classifier, which takes a high-resolution image or a generated SR image as input and evaluates the likelihood of the input image being produced by the generator. During ESRGAN training, the discriminator is initially trained by minimizing the adversarial loss between itself and the generator. Subsequently, the generator is trained with fixed discriminator parameters, minimizing the content-aware loss between the generated image and the actual high-resolution image.
 
While ESRGAN has been proven effective in enhancing the detail and clarity of natural images, directly applying it to restore EDs may not be reasonable. On the one hand, EDs primarily comprise simple geometric symbols, lines, and text annotations, with limited contextual information compared to the richer context found in natural images. This scarcity of contextual information poses challenges for ESRGAN, potentially leading to distortions and over-enhancement. On the other hand, the shape and location of graphical elements on EDs generally bear specific physical meanings, requiring higher SR precision than natural images. Severe distortions and excrescent information introduced by SR may adversely impact the interpretation and analysis of EDs, warranting the research on a modified version of ESRGAN dedicated to processing low-quality EDs.

\subsection{Faster R-CNN for Graphical symbol Detection}
\label{sub:faster_r_cnn_for_graphic_symbol_detection}

\begin{figure}[!htb]
  \centering
  \includegraphics[width=4in]{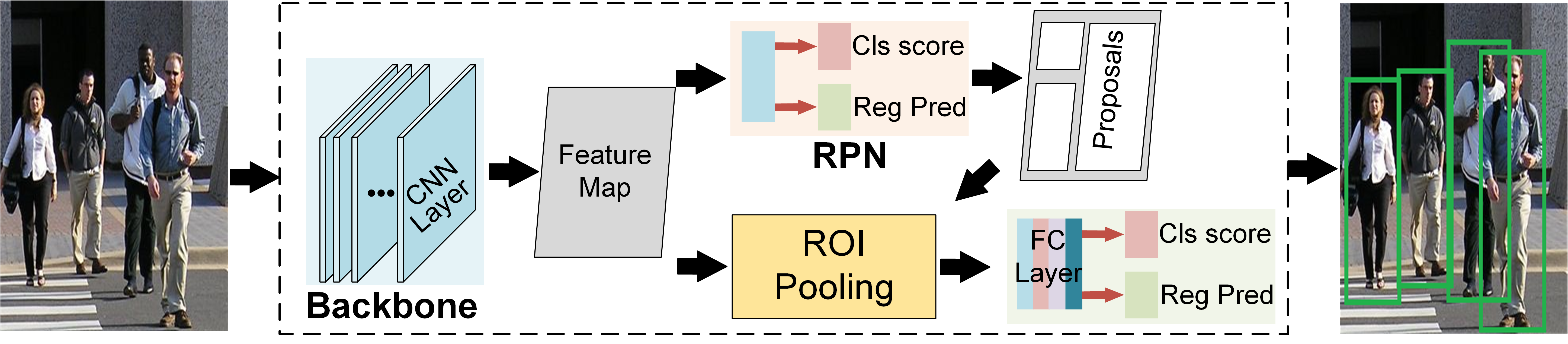}
  \caption{The network structure of Faster R-CNN.}
  \label{fig:faster_rcnn}
\end{figure}

Faster R-CNN \cite{ren2015faster} is a widely employed object detection model, initially designed for natural images, and subsequently extended for various industrial applications. In the proposed framework, this model is utilized to detect graphical symbols on EDs. The primary components of Faster R-CNN, illustrated in \figref{fig:faster_rcnn}, include a convolutional feature extraction network (backbone), a region proposal network (RPN), and a region classification network. Before inputting an image into the Faster R-CNN model, the image is rescaled to a predefined fixed size. Upon receiving the image, the backbone network extracts feature maps. The RPN generates proposals based on these feature maps, and the region classification network performs bounding box regression and classification for each proposal. The Faster R-CNN model adopts an end-to-end training approach, wherein the backbone network, RPN, and region classification network are jointly trained using a multi-task loss function. This training methodology allows for seamless optimization of the entire model to empower its object detection performance.

While Faster R-CNN performs effectively on natural images, challenges arise when applying it to EDs. Graphical symbols in EDs are typically much smaller compared to the overall size of the drawings. The original Faster R-CNN model may struggle to detect small objects on large images. Furthermore, the effectiveness of Faster R-CNN heavily relies on the quality of the input images. If the input EDs exhibit quality degradation, the detection accuracy may further deteriorate. These limitations underscore the need for refining the model to ensure accurate and reliable detection of graphical symbols on low-quality EDs.

\section{Low-Quality ED Recognition Framework}
\label{sec:low_quality_ed_recognition_framework}

This section provides a comprehensive description of the proposed low-quality ED recognition framework. It begins by introducing the overall architecture of the framework and proceeds to discuss the constituent modules, including image preprocessing, restoration, and symbol detection, in detail. The multi-stage task-driven collaborative learning strategy used to jointly train the modified ESRGAN and Faster R-CNN models is also thoroughly explained, with the benefits of this synergistic training approach in enhancing the overall graphical symbol recognition performance finally emphasized.

\subsection{Overall Architecture of Proposed Framework}
\label{sub:overall_architecture_of_proposed_framework}

\begin{figure}[!htb]
  \centering
  \includegraphics[width=2.8in]{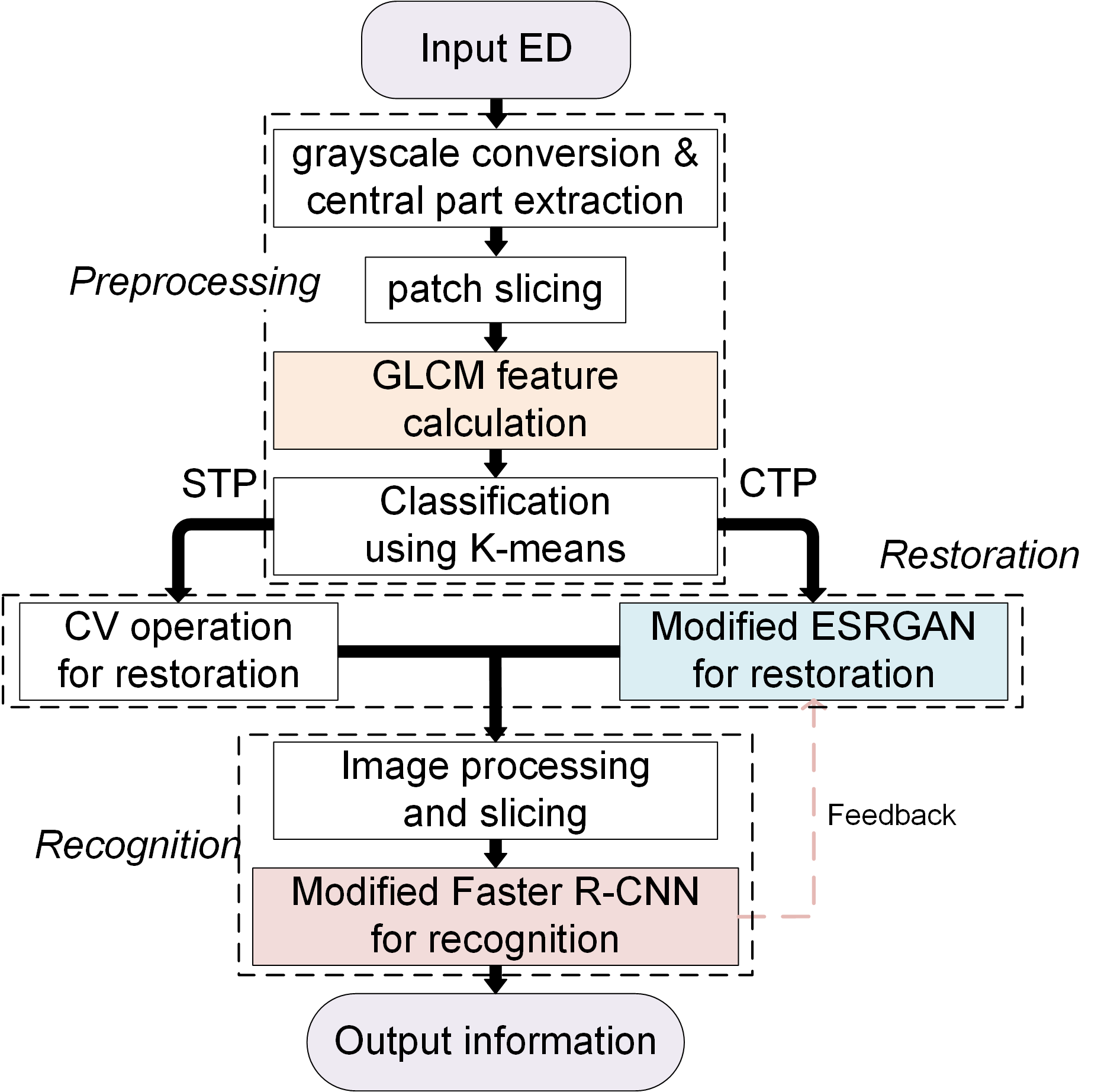}
  \caption{The overall architecture of the proposed low-quality ED recognition framework.}
  \label{fig:overall_workflow}
\end{figure}

The overall architecture of the proposed framework for low-quality ED recognition is depicted in Fig. \ref{fig:overall_workflow}. Given a low-quality ED, image preprocessing is first performed to convert it to a grayscale version. The grayscale ED is then sliced into several patches. The GLCM features are computed for these patches, which are fed into a K-means clustering model to divide the patches into two classes, STPs and CTPs. To deal with the degradation of the input ED, the two classes of patches are sent to different modules for restoration. STPs are restored using some simple CV heuristics, while CTPs are restored using a modified ESRGAN model. The restored patches are reorganized into a high-quality ED. After another round of processing and slicing, the resulting high-quality patches are finally sent to a modified Faster R-CNN model for symbol detection and recognition.

\subsection{Preprocessing of Low-Quality EDs}
\label{sub:preprocessing_of_low_quality_eds}

The preprocessing module takes low-quality EDs as input and performs several operations in sequence, including grayscale conversion, central part extraction, patch slicing, GLCM feature calculation, and K-means clustering. The details of these preprocessing operations are described as follows.

\subsubsection{Grayscale Conversion and Central Part Extraction} 
\label{ssub:grayscale_conversion_and_central_part_extraction}

At the very beginning of the preprocessing module, grayscale conversion is performed on the input ED. This operation transforms a color image represented by the red, green, and blue (RGB) channels into a single-channel grayscale image by calculating the weighted average of the RGB channels. All subsequent processing is carried out on the grayscale image. 

To avoid meaningless analysis of the blank padding of the ED, central part extraction is then performed by applying an edge detection algorithm to the input ED and identifying the minimal bounding box that envelops its main body. It is worth noting that due to the potential low quality of the input ED, optional noise reduction, binarization, and sharpening operations may be performed before edge detection to improve the robustness of central part extraction.

\subsubsection{Patch Slicing} 
\label{ssub:patch_slicing}

\begin{figure}[!htb]
  \centering
  \includegraphics{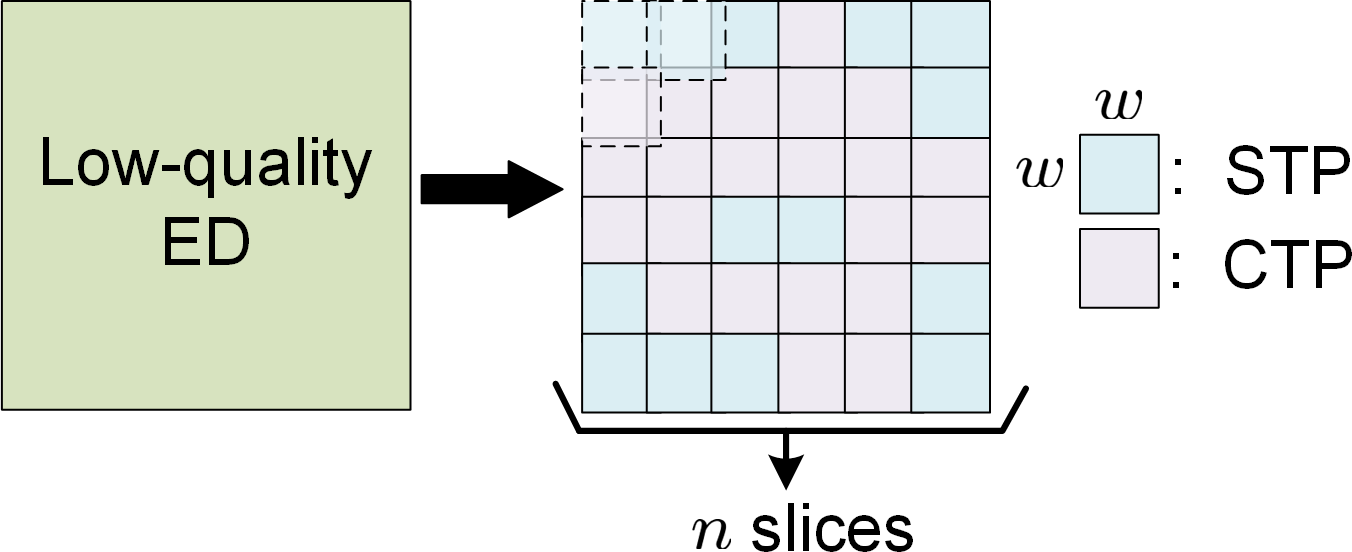}
  \caption{Diagram for the slicing of ED.}
  \label{fig:Diagram for the slicing of ED}
\end{figure}

Subject to the network structure, the input image size of a deep learning-based CV model is often fixed and small. Although the sizes of different types of EDs may vary considerably, they are basically much larger than the required input size. If they were directly scaled to suit the input size, it would cause further damage to the ED and lead to performance degradation for subsequent recognition. Therefore, a patch slicing process, as shown in \figref{fig:Diagram for the slicing of ED}, is performed to cut a complete ED into $n$ patches with a size of $w \times w$. Each patch corresponds to a part of the original ED image and overlaps with an adjacent patch by $p$ pixels in the vertical or horizontal direction.

\subsubsection{GLCM Feature Calculation and Patch Classification} 
\label{ssub:glcm_feature_calculation_and_patch_classification}

In \cite{kong2021classsr}, Kong \textit{et al.} demonstrate that different parts of an image present different levels of difficulty in restoration. Specifically, areas with complex textures are harder to restore and require a more sophisticated SR model. In contrast, patches with simple textures can be well restored with basic CV heuristics. Even after extracting the central part of an ED, the remaining part may still contain many blank areas or sparsely located geometric shapes. Using a complex SR model to recover all ED patches is overkill and greatly reduces the efficiency of the framework. Therefore, the proposed framework computes the four GLCM features discussed in Section \ref{sub:glcm_for_grayscale_texture_characterization}, i.e., the dissimilarity $M_d$, homogeneity $M_h$, energy $M_e$, and entropy $M_p$, to measure the texture complexity of each patch. To avoid the interference of noise pixels on the calculation of GLCM features, binarization is performed in advance to keep the grayscale value of the background consistent among different patches.

Once the GLCM features of patches are calculated, a K-means clustering model is used to classify them into two classes: CTP and STP. CTP contains one or more complete graphical symbols, while STP only contains simple geometric shapes or fragmented graphical symbols, with a lot of space being empty. K-means is an unsupervised learning model that aims to divide a set of data points into $k$ distinct clusters such that the similarity of data points within a cluster is maximized, and the similarity between clusters is minimized. K-means clustering is computationally efficient and highly adaptable to numerically characterized data. The concrete procedure of GLCM feature calculation and patch classification is shown in Algorithm \ref{alg:glcm_feature_calculation_and_patch_classification}.

\begin{algorithm}[!htb]
\caption{GLCM Feature Calculation and Patch Classification}
\label{alg:glcm_feature_calculation_and_patch_classification}
  \SetAlgoLined
  \DontPrintSemicolon
  \KwIn{The set of all patches $\mathcal{C}$, tunable parameters $\alpha_1, \alpha_2, \alpha_3, \alpha_4$}
  \KwOut{The sets of CTP $\mathcal{C}_{CTP}$ and STP $\mathcal{C}_{STP}$}
  \;
  Initialize: $\mathcal{C}_{CTP} \leftarrow \varnothing$, $\mathcal{C}_{STP} \leftarrow \varnothing$, GLCM feature set $\mathcal{I} \leftarrow \varnothing$, $k=2$\;
  \ForEach{$c \in \mathcal{C}$}{
    Compute the GLCM $G_{\Delta x, \Delta y}$ for $c$ according to \eqref{eq:glcm}\;
    Compute the statistical measures $M_d$, $M_h$, $M_e$, $M_p$ according to \eqref{eq:dissimilarity}--\eqref{eq:entropy} as the GLCM features of $c$\;
    Vectorize the GCLM features as $\bm{i}_c = (\alpha_1 M_d, \alpha_2 M_h, \alpha_3 M_e, \alpha_4 M_p)$\;
    $\mathcal{I} \leftarrow \mathcal{I} \cup \{\bm{i}_c\}$
  }
  Randomly select $k=2$ samples from $\mathcal{I}$ as the barycenters ${\bm{u}_1, \bm{u}_2}$\;
  Initialize: $\tilde{\bm{u}}_1 = \tilde{\bm{u}}_2 = (0,0,0,0)$\;
  \While{$(\tilde{\bm{u}}_1 \neq \bm{u}_1) \land (\tilde{\bm{u}}_2 \neq \bm{u}_2)$}{
   \uIf{$|\mathcal{C}_{CTP}| > 0 \land |\mathcal{C}_{STP}| > 0$}{
        $\bm{u}_1\leftarrow\tilde{\bm{u}}_1$, $\bm{u}_2\leftarrow\tilde{\bm{u}}_2$\;
   }
   \ForEach{$c \in \mathcal{C}$}{
        $d_{i1} = \| \bm{i}_c - \bm{u}_1 \|^2$, $d_{i2} = \| \bm{i}_c - \bm{u}_2 \|^2$\;
        \uIf{$d_{i1} \leq d_{i2}$}{$\mathcal{C}_{STP} \Leftarrow \mathcal{C}_{STP} \cup \{c\}$\;}
        \Else{$\mathcal{C}_{CTP} \Leftarrow \mathcal{C}_{CTP} \cup \{c\}$\;}
   }
   $\tilde{\bm{u}}_1 \leftarrow \frac{1}{|\mathcal{C}_{STP}|}\sum_{c' \in \mathcal{C}_{STP}} \bm{i}_{c'}$, $\tilde{\bm{u}}_2 \leftarrow \frac{1}{|\mathcal{C}_{CTP}|}\sum_{c' \in \mathcal{C}_{CTP}} \bm{i}_{c'}$\;
  }
  \KwRet{$\mathcal{C}_{CTP}$, $\mathcal{C}_{STP}$}
\end{algorithm}

\subsection{Restoration of ED Patches}
\label{sub:restoration_of_ed_patches}

\begin{figure}[!htb]
  \centering
  \includegraphics{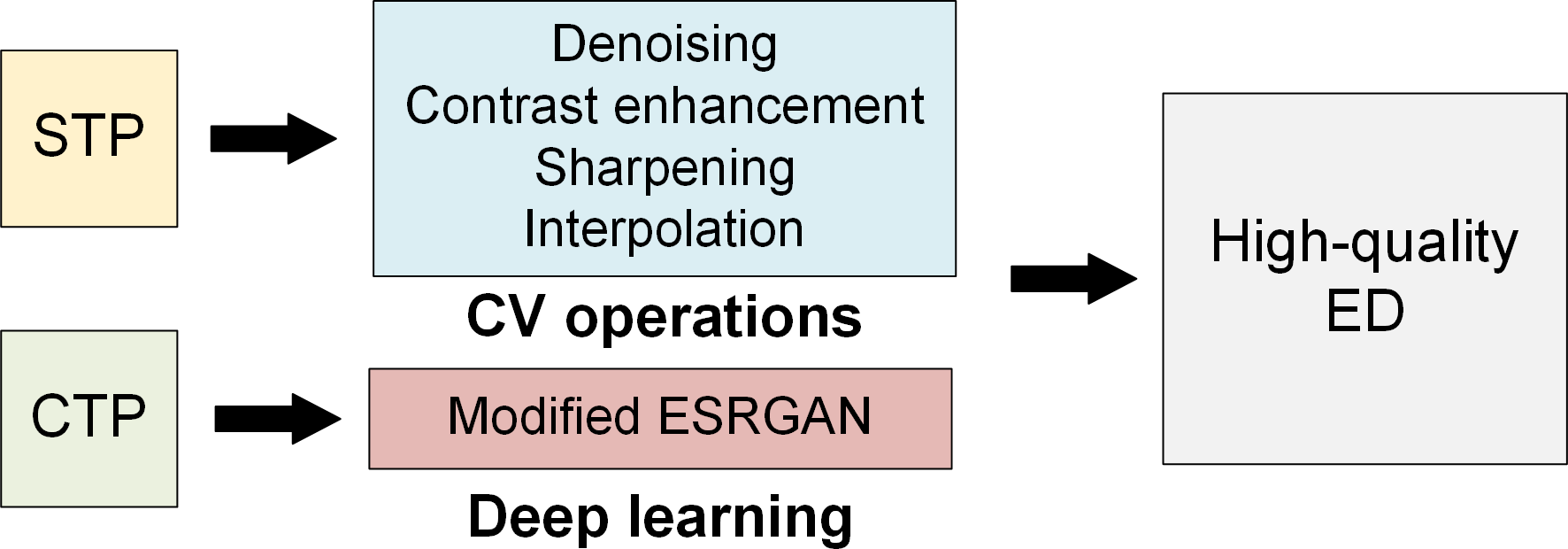}
  \caption{The workflow of the restoration module.}
  \label{fig:The workflow diagram of the restoration module}
\end{figure}

The preprocessing module in the proposed framework is followed by a restoration module, which aims to improve the definition and clarity of the preprocessed low-quality ED patches. The restoration module is tailored to the two types of patches, with different techniques applied depending on the difficulty of restoration. A schematic diagram of the resolution module is shown in the \figref{fig:The workflow diagram of the restoration module}.

\subsubsection{Restoration of STP}
\label{ssub:restoration_of_stp}

Practical EDs often contain a substantial number of STPs. If all STPs were enhanced using deep learning models, it would consume a significant amount of resources and time. As a matter of fact, most STPs only contain some simple geometric shapes, such as line segments or parts of graphical symbols, leaving the majority of the canvas blank. As a result, restoring STPs using basic CV heuristics is acceptable and rarely affects the subsequent symbol recognition task. In this context, denoising, contrast enhancement, sharpening, and interpolation operations are applied to restore STPs in the proposed framework, in order to achieve a balance between restoration quality and computational cost.

\begin{figure}[!htb]
  \centering
  \includegraphics{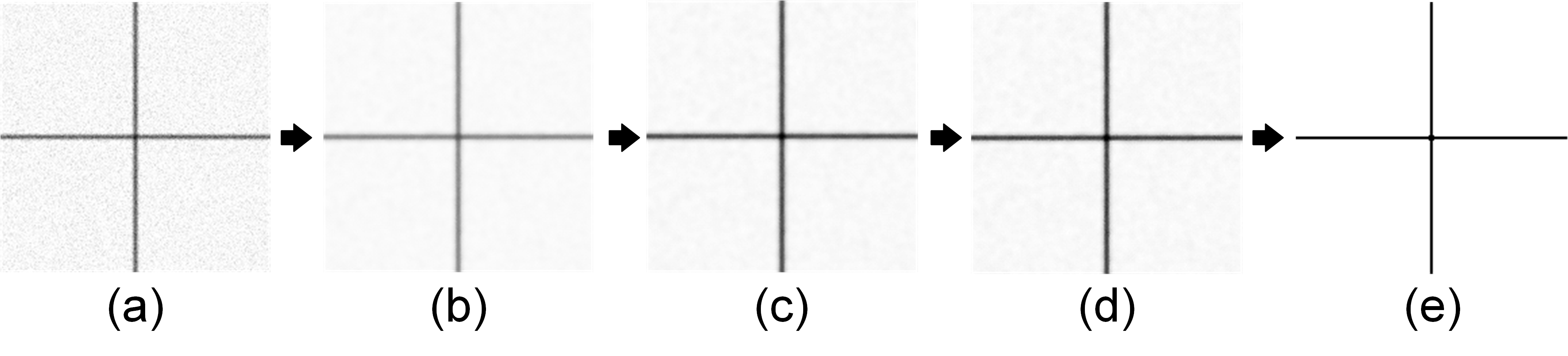}
  \caption{The restoration process of an STP using basic CV heuristics.}
  \label{fig:Schematic diagram of the recovery process using a series of plain vision algorithms}
\end{figure}

The STP restoration process is illustrated in \figref{fig:Schematic diagram of the recovery process using a series of plain vision algorithms}. The original STP containing a line crossing is shown in (a). First, a bilateral filter is applied to (a), resulting in a smoothed but denoised version shown in (b). Next, the contrast stretching algorithm is applied to (b) to enhance its contrast, yielding (c). A Laplacian of Gaussian filter is then used to sharpen the STP, producing (d). Finally, edge-based interpolation is performed on (d), resulting in a restored version of (a), shown in (e).

\subsubsection{Restoration of CTP}
\label{ssub:restoration_of_ctp}

In contrast to STPs, CTPs typically contain a wealth of contextual information. Using CV heuristics to restore CTPs may lead to information loss, distortion, and undesirable artifacts and detriments to the subsequent symbol recognition task. Therefore, an ESRGAN-based deep learning model is used in the proposed framework to restore CTPs. While ESRGAN has a strong restoration capability when dealing with natural images, it is fully suitable for processing EDs due to the differences in their graphical patterns. Thus, the original ESRGAN model is modified in three ways to enhance its performance toward CTPs:
\begin{enumerate}
    \item This paper reduces the depth of the original ESRGAN model from 23 RRDB blocks to 16. The original ESRGAN model, designed for natural images, uses a very deep architecture to learn the intricate mapping between low-resolution and high-resolution images and ensure high SR performance under various scenarios. However, graphical symbols in EDs have a more prominent foreground, making it unnecessary to use such a deep model. Additionally, reducing the depth of the network can also decrease computation and reduce the number of required training samples, which is beneficial for practical applications in engineering fields.
    \item This paper replaces the regular Leaky Rectified Linear Unit (LReLU) activation function in the original ESRGAN with the Randomized Leaky Rectified Linear Unit (RReLU) activation function \cite{xu2015empirical}. The RReLU function is expressed as
    \begin{equation}
        \mathrm{RReLU}(x)= \begin{cases}
            x & \text{if $x \geq 0$} \\ 
            ax & \text{if $x < 0$}
        \end{cases}
        \label{eq:rrelu}
    \end{equation}
    where $a$ is randomly sampled from a uniform distribution $\mathcal{U}(0.1,0.4)$. The random slope of the negative values in RReLU provides the model a with some regularization effects and thus improves its generalizability.
    \item This paper incorporates the gradient loss proposed in \cite{wang2020scene} into the loss function of the ESRGAN generator. Originally used in text image SR to sharpen the boundaries of restored characters, the gradient loss can be expressed as
    \begin{equation}
        L_{gp} = \mathbb{E}_x \|\nabla I(x^{hq}) - \nabla I(x^{res}) \|_1
        \label{eq:gradient_loss}
    \end{equation}
    where $\mathbb{E}_x$ denotes the expectation over every pair of low-quality and high-quality EDs $(x^{lq}, x^{hq})$ in the training dataset, $\nabla I(\cdot)$ returns the gradient field of an input image, $x^{res}=G(x^{lq},\theta)$ is the restored ED output by the generator $G$ based on the input low-quality ED $x^{lq}$. Since graphical symbols on EDs typically contrast strongly against the background, incorporating the gradient loss can hopefully improve the clarity of the restored ED. 
\end{enumerate}

The training of the modified ESRGAN model is similar to the training of the original ESRGAN model, i.e., alternatively training the discriminator $D$ and the generator $G$. The discriminator loss 
\begin{equation}
    L_D=-\mathbb{E}_{x^{hq}}[\log(D_r(x^{hq}, x^{res}))] - \mathbb{E}_{x^{res}}[\log(1 - D_r(x^{res}, x^{hq}))]
    \label{eq:discriminator_loss}
\end{equation}
where $D_r(x, y)=\sigma(D(x)-\mathbb{E}_y[D(y)])$ is the relativistic average discriminator function, with $\sigma(\cdot)$ the activation function and $D(\cdot)$ the non-transformed discriminator output. When the restored ED $x^{res}$ is similar to the high-quality ED $x^{hq}$, $D(x^{hq}, x^{res})\approx 1$ and $D(x^{res}, x^{hq})\approx 0$. 

With the gradient loss $L_{gp}$ taken into account, the total generator loss
\begin{equation}
    L_G = L_p + \alpha L_1 + \beta L_{gp} + \gamma L_a
\label{eq:generator_loss}
\end{equation}
where $L_p=\mathbb{E}_x\|F(x^{hq})-F(x^{res})\|_1$ is the perceptual loss calculating the difference between the feature maps of the high-quality ED $x^{hq}$ and the restored ED $x^{res}$. $L_1=\mathbb{E}_x\|x^{hq}-x^{res}\|_1$ is the content loss measuring the difference between $x^{hq}$ and $x^{res}$. $L_a$ is the adversarial loss symmetrical to the discriminator loss, which is computed as
\begin{equation}
    L_a=-\mathbb{E}_{x^{hq}}[\log(1 - D_r(x^{hq}, x^{res}))] - \mathbb{E}_{x^{res}}[\log(D_r(x^{res}, x^{hq}))]
\label{eq:adversarial_loss}
\end{equation}
The coefficients $\alpha$, $\beta$, and $\gamma$ are weight parameters balancing the four loss terms. 

\subsection{Graphical Symbol Recognition}
\label{sub:graphical_symbol_recognition}

After restoration, the output quality-enhanced STPs and CTPs are merged to form an ED image with the same size as the input ED. The restored ED is then sliced into overlapping patches again according to the description in Section \ref{ssub:patch_slicing}, with patches having a larger size than the previous slicing operation to ensure that each graphical symbol appears fully on at least one patch. These patches are then input into the next module, where a modified Faster R-CNN model is used to determine the location and type of graphical symbols. Although Faster R-CNN is widely used as an object detection model in industrial applications, it still faces some challenges in identifying graphical symbols on EDs. This is mainly because graphical symbols may appear in varying sizes, with some being very large and others being quite small relative to the patch size. To improve the recognition accuracy of the original Faster R-CNN, the proposed framework modifies its network structure in two aspects.

\begin{figure}[!htb]
  \centering
  \includegraphics[width=3.3in]{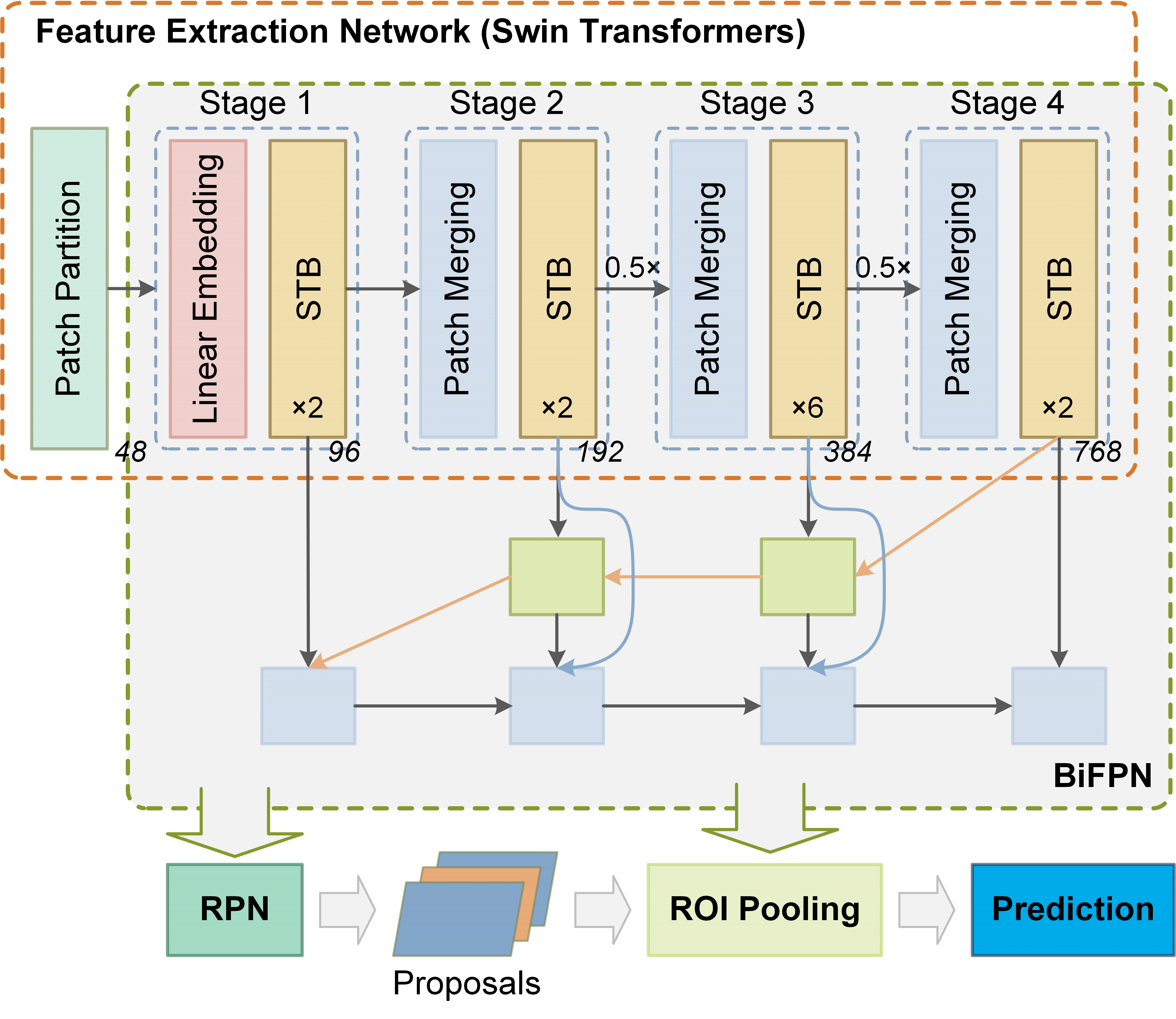}
  \caption{The network structure of the modified Faster R-CNN model.}
  \label{fig:The network structure of Modified Faster R-CNN}
\end{figure}

\begin{enumerate}
    \item As depicted in \figref{fig:The network structure of Modified Faster R-CNN}, the CNN-based backbone of the original Faster R-CNN model is replaced with the Swin Transformer \cite{liu2021swin}. The Swin Transformer excels in capturing the global context of large images because the self-attention mechanism involved in the Swin Transformer Block (STB) is performed not only within a single local window but also across multiple distant windows. Thanks to the advanced feature representation capability provided by the Swing Transformer, the modified Faster R-CNN model is expected to achieve good symbol recognition performance, even with relatively large input patches. This also has the side benefit of improving the computational efficiency.
    \item The layers of the original Faster R-CNN model is reorganized into a Bi-directional Feature Pyramid Network (BiFPN) \cite{tan2020efficientdet}  to improve the multi-scale feature extraction capability to adapt to the varying size of graphical symbols on EDs. Unlike the previously used Feature Pyramid Network (FPN), BiFPN introduces additional connections from bottom to top to enhance information mobility, enabling bi-directional feature propagation and avoiding the information loss caused by the top-down unidirectional information flow limitation of FPN. Additionally, BiFPN assigns a dynamic weight to each feature layer, which also benefits the detection of small symbols without sacrificing efficiency.
\end{enumerate}

The modified Faster R-CNN inherits the training method of the original Faster R-CNN, which involves a four-step alternating process to train the Region Proposal Network (RPN) and the Fast R-CNN backbone \cite{ren2015faster}. Accordingly, the overall loss function is divided into two parts: the RPN loss, denoted as $L_{RPN}$, and the Fast R-CNN loss, denoted as $L_{task}$. Both $L_{RPN}$ and $L_{task}$ have a simiar form, which is a weighted sum of classification loss and bounding-box regression loss. Taking the Fast R-CNN loss $L_{task}$ as an example, the loss function can be expressed as
\begin{equation}
    L_{task}=\mathbb{E}_x [L_{cls}(y, \hat{y}(x))] + \mu \mathbb{E}_x [p_x^* L_{reg}(t, \hat{t}(x))] 
    \label{eq:fast_rcnn_loss}
\end{equation}
Here, $y$ and $\hat{y}(x)$ are the ground-truth class and the predicted class of an input image $x$, respectively. $t$ and $\hat{t}(x)$ are vectors that represent the actual and predicted offsets between the real location of the object in $x$ and the proposal given by RPN. Moreover, $p_x^*$ represents whether an object exists in the proposal, i.e., $p_x^*=1$ if an object is in the box and $p^*=0$ otherwise. The classification loss $L_{cls}(\cdot,\cdot)$ utilizes the multi-class cross-entropy as the underlying loss function, while the regression loss $L_{reg}(\cdot,\cdot)$ employs the smooth L1-loss as the underlying loss function.

\subsection{Multi-Stage Task-Driven Collaborative Training} 
\label{sub:multi_stage_task_driven_collaborative_training}

The low-quality ED recognition framework proposed in this paper involves two deep learning models, namely the modified ESRGAN model and the modified Faster R-CNN model. If these two models are trained independently, the SR model may introduce artifacts or distortions to the restored ED. Although these distortions may not affect human comprehension, they could still adversely impact the performance of the downstream graphical symbol detection model, potentially leading to false detection or misidentification. Collaborative learning is an effective way to improve the overall performance and generalizability of composite models \cite{bai2018sod,wang2022remote,shim2022road}. To overcome the drawback of independent training, this paper proposes a multi-stage, task-driven collaborative training strategy to jointly train the two deep learning models.

\textbf{Stage 1 (Pre-training):} The modified ESRGAN model and the modified Faster R-CNN model are pre-trained individually according to the methods discussed in Section \ref{ssub:restoration_of_ctp} and Section \ref{sub:graphical_symbol_recognition}, respectively. After that, the pre-trained discriminator of the modified ESRGAN model is discarded, while the generator is concatenated with the pre-trained modified Faster R-CNN model to form a learning pipeline. 

\textbf{Stage 2 (Fine-tuning of the modified ESRGAN generator):} The parameters of the modified Faster R-CNN model are frozen, with the generator network fine-tuned based on the following loss function
\begin{equation}
    L_{gen}^{ft} = \lambda_1 L_G'(x^{hq}, x^{res}) + \lambda_2 L_{task}(x^{res}, y, t)
\label{eq:generator_update_loss}
\end{equation}
where $L_G' = L_p + \alpha L_1 + \beta L_{gp}$, i.e., the generator loss in \eqref{eq:generator_loss} excluding the adversarial loss. Considering that $x^{res}=G(x^{lq},\theta)$, given a matched group of low-quality ED patch $x^{lq}$, high-quality ED patch $x^{hq}$, symbol class label $y$, and symbol location offset $t$, the parameters of the generator network $\theta$ can be effectively optimized by applying stochastic gradient descent to the back-propagated error.

\textbf{State 3 (Fine-tuning of the modified Faster R-CNN}: The parameters of the modified ESRGAN model are frozen, with the modified Faster R-CNN model fine-tuned using the last two steps of the four-step alternating training method \cite{ren2015faster}.

The proposed multi-stage task-driven collaborative learning strategy offers several benefits:
\begin{enumerate}
  \item It prevents the two deep learning models from becoming a simple splice with no connection in between, which hinders the optimization of the overall low-quality ED recognition accuracy.
  \item It allows the low-quality ED to be restored in a task-driven manner, i.e., contributing most to the downstream graphical symbol recognition task rather than the human visual perception.
  \item It avoids the slow or even impossible convergence caused by the entire network finding dynamic equilibrium between the two constituent models and accelerates the overall training process.
\end{enumerate}

\subsection{Training Data Preparation}
\label{sub:training_data_preparation}

As discussed in Section \ref{sub:multi_stage_task_driven_collaborative_training}, each training sample in the proposed framework is a 4-tuple, comprising a low-quality ED patch, the corresponding high-quality ED patch, as well as the class labels and location offsets of graphical symbols on the patch. However, it is often challenging to collect a sufficient number of matched pairs of low-quality and high-quality EDs in practice, which hampers the practical application of the proposed framework. To overcome this challenge, many image restoration studies in the literature generate low-quality images from high-quality images using the classic quality degradation model
\begin{equation}
    \tilde{x}^{lq}=U(x^{hq})=C_{JPEG}[S(y \otimes k, r)+n]
    \label{eq:quality_degradation_model}
\end{equation}
where $U(\cdot)$ represents the degradation function, $\otimes$ the convolution operation, $k$ the fuzzy kernel, $S(\cdot, r)$ the downsampling operation with a ratio of $r$, $n$ the added noise, and $C_{JPEG}(\cdot)$ the JPEG compression operation. According to \eqref{eq:quality_degradation_model}, a low-quality image $\tilde{x}^{lq}$ can be derived from its high-quality counterpart $x^{hq}$.

However, the low-quality images created using this classic quality degradation model tend to be overly homogeneous, which may reduce the generalizability of the restoration model and lead to poor performance of downstream models \cite{pei2019effects}. To enhance the flexibility of the classic quality degradation model, a multi-order degradation model is proposed in \cite{wang2021real}, which is formulated as
\begin{equation}
    \tilde{x}^{lq}=U^n(x^{hq})=U_n (U_{n-1}(\dots U_1(x^{hq})\dots))
\end{equation}
where $n$ denotes the number of orders. Besides, a more practical way of synthesizing quality-degraded images is proposed in \cite{zhang2021designing}. Inspired by these approaches, this paper proposes a new method for generating training samples of low-quality EDs, taking into account the various degradation effects commonly seen in real-world low-quality EDs. 

\begin{figure}[!htb]
  \centering
  \includegraphics[width=3.8in]{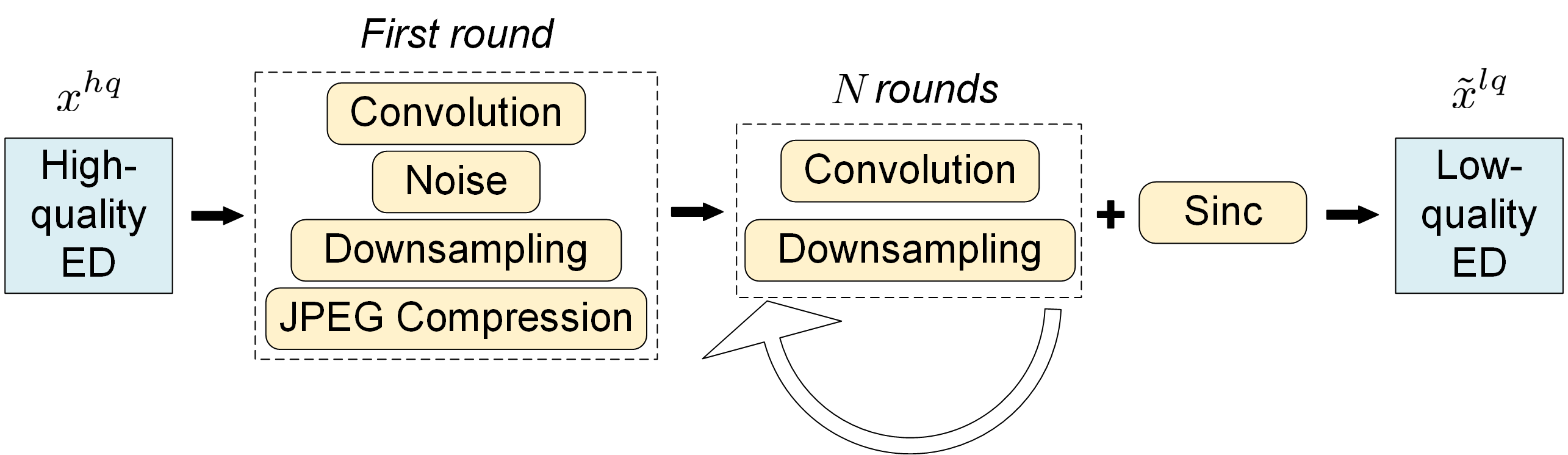}
  \caption{The flow chart of the proposed ED quality degradation method.}
  \label{fig:Flow chart of low-quality ED data generation}
\end{figure}

As illustrated in \figref{fig:Flow chart of low-quality ED data generation}, a high-quality ED $x^{hq}$ first undergoes a round of classic degradation as per \eqref{eq:quality_degradation_model}. Notably, the four operations in the degradation model need not necessarily be performed. Their application and parameters can be randomly determined, as long as their execution sequence is consistent with \eqref{eq:quality_degradation_model}. After the first round of degradation, $N$ rounds of degradation is then performed recursively, with each round of degradation containing only convolution-based blurring and downsampling, as these two are the most common degradation effects on practical EDs. Finally, a sinc filter is applied to simulate the ringing artifacts. Based on the proposed ED quality degradation method, multiple low-quality EDs with different quality degradation effects can be generated from the same high-quality ED. This is beneficial for training the proposed framework to adapt to the diverse degradation patterns in the real world. 

\section{Experiments and Discussions}
\label{sec:experiments_and_discussions}

This section evaluates the performance of the proposed framework using real-world low-quality substation single-line diagrams (SLDs), a common type of ED in electric power engineering. The SLDs used in the evaluation are provided by our cooperative partners from power utilities. In addition, the proposed framework has been integrated into RelayCAC, a power system protective relay setting coordination software widely used in China. The implementation details of this integration are also presented in this section.

\subsection{Experimental Setup} 
\label{sub:experimental_setup}

SLDs represent electrical components using graphical symbols and their connections by lines. The SLDs used in the experiments are of varying sizes, with the average size of high-quality SLDs being approximately $5800 \times 4100$ pixels and the average size of low-quality SLDs being approximately $1000 \times 650$ pixels. Eight typical symbols, including disconnectors (DCR), circuit breakers (BKR), ground lead disconnectors (GLD), power transformers (TFM), inductors (IND), capacitors (CAP), generators (GEN), and grounding devices (GND), are selected as the classes of interest in the recognition performance evaluation. 

The experiments are conducted on a server with two Intel Xeon Gold 6248R 3.00 GHz CPUs, 128 GB RAM, and an NVIDIA RTX 3090 GPU. The operating system is Ubuntu Linux 22.04, and the Python interpreter version is 3.8. The deep learning models are implemented based on the Pytorch 2.1.1 library with CUDA Toolkit 11.8 and the MMdetection toolbox \cite{mmdetection}. 

At the beginning of the experiments, low-quality SLDs are preprocessed according to the method described in Section \ref{sec:preliminary_technology}. The four tunable parameters in Algorithm \ref{alg:glcm_feature_calculation_and_patch_classification} are set as $\alpha_1=0.3$, $\alpha_2=0.1$, $\alpha_3=0.2$, $\alpha_4=0.4$, respectively. When training the framework based on the multi-stage task-driven collaborative strategy discussed in Section \ref{sub:multi_stage_task_driven_collaborative_training}, the modified ESRGAN is trained for 1,500 iterations in the first stage, with a batch size of 16 and an initial learning rate of $2 \times 10^{-4}$, and for 2,000 iterations in the second stage, with a batch size of 8 and an initial learning rate of $1 \times 10^{-4}$. The three coefficients in the generator loss \eqref{eq:generator_loss} are set as $\alpha=1\times 10^{-2}$, $\beta=1\times 10^{-4}$, and $\gamma=5\times 10^{-3}$, while the weighting coefficients in the fine-tuning loss \eqref{eq:generator_update_loss} are set as $\lambda_1=\lambda_2=1$. For the modified Faster R-CNN model, it is trained for 3,200 iterations in the first stage, with a batch size of 4 and an initial learning rate of $1 \times 10^{-4}$, and for 2,400 iterations in the third stage, with the batch size and the initial learning rate remaining unchanged. The AdamW algorithm is used as the underlying optimizer of the training process, with beta parameters set to $\beta_1=0.9$ and $\beta_2=0.999$. The patch size for image restoration is set to $200\times 200$, and the patch size for graphical symbol recognition is set to $1000 \times 1000$. The magnification factor used in image restoration is set to 4. The Intersection over Union (IoU) threshold used in symbol recognition is set to 0.9. Among all bounding box predictions corresponding to a specific target, the one with the largest IoU is selected as the final detection result. Precision, recall and F1-score are used as the performance metrics for evaluating the proposed framework.

\subsection{Evaluation of Overall Performance} 
\label{sub:evaluation_of_overall_performance}

\begin{figure}[!htb]
  \centering
  \includegraphics[width=4in]{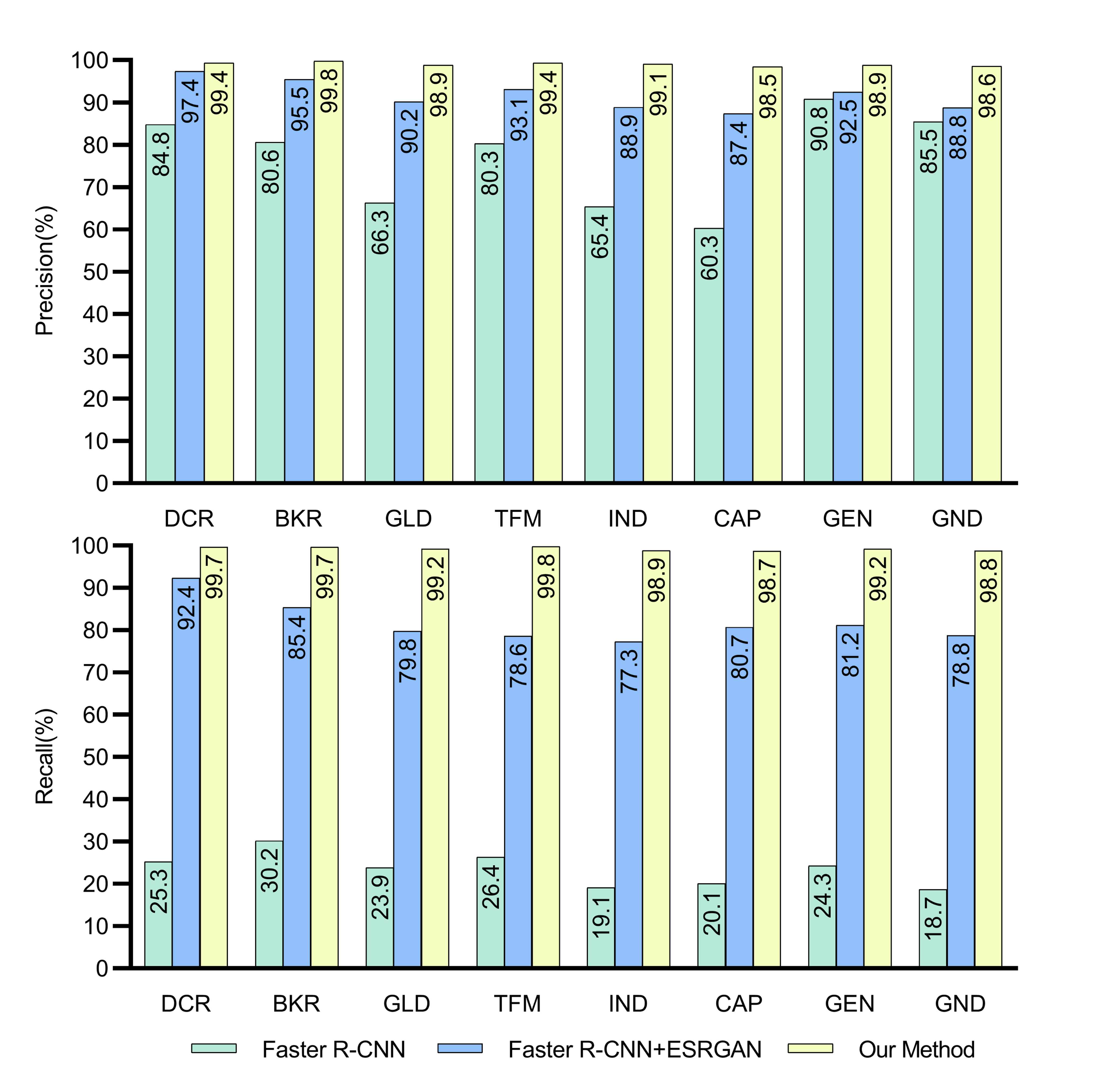}
  \caption{Graphical symbol recognition results of the three tested models.}
  \label{fig:Graphic symbols recognition results for three situations}
\end{figure}

The first experiment aims to verify the overall low-quality ED recognition performance of the proposed framework. Three different models, including a single Faster R-CNN model, a combination of independently trained ESRGAN and Faster R-CNN models, and the proposed framework, are used to identify the graphical symbols on the low-quality SLDs, which form an ablation study. The preprocessing and the hyperparameters are kept consistent for the three models. The low-quality EDs used in the experiment are obtained using the degradation model discussed in Section \ref{sub:training_data_preparation}, with a maximum degradation order of 5. The recognition results for different classes of graphical symbols are shown in \figref{fig:Graphic symbols recognition results for three situations}. 

The results show that using a single Faster R-CNN model to recognize low-quality SLDs leads to poor performance, with a low recall rate indicating that numerous graphical symbols are undetected. This is likely due to the significant distortion of symbol characteristics and the presence of noise. The introduction of the ESRGAN model considerably improves the precision and recall, highlighting the contribution of restoring low-quality EDs to the downstream symbol recognition task. The proposed framework outperforms the independently trained ESRGAN and Faster R-CNN models in terms of all eight classes of symbols of interest. The lowest precision and recall of the proposed framework achieve 98.6\% and 98.7\%, respectively, demonstrating its powerful recognition performance towards low-quality EDs with varying degrees of degradation. Furthermore, the improvement over independently trained models confirms the effectiveness of the proposed multi-stage task-driven collaborative training strategy in fine-tuning the image restoration model to prioritize machine vision over human vision.

\begin{figure}[!htb]
  \centering
  \includegraphics[width=4.2in]{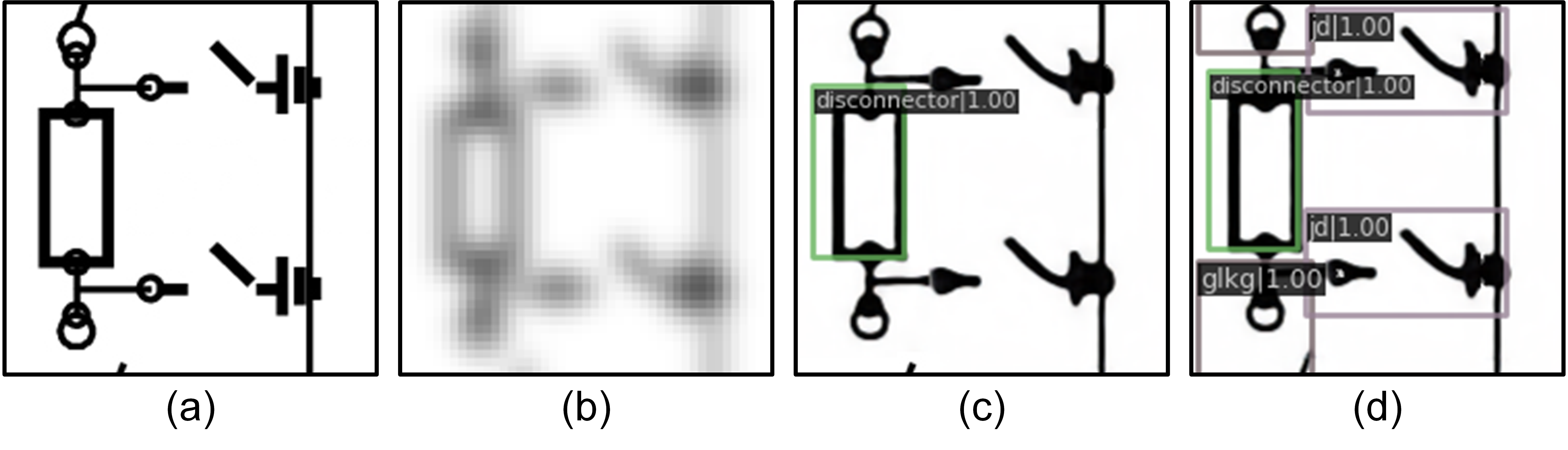}
  \caption{An intuitive comparison of different models using a severely degraded SLD patch.}
  \label{fig:Detection performance of three different methods in the face of heavily degraded drawings}
\end{figure}

\begin{figure}[!htb]
  \centering
  \includegraphics[width=\textwidth]{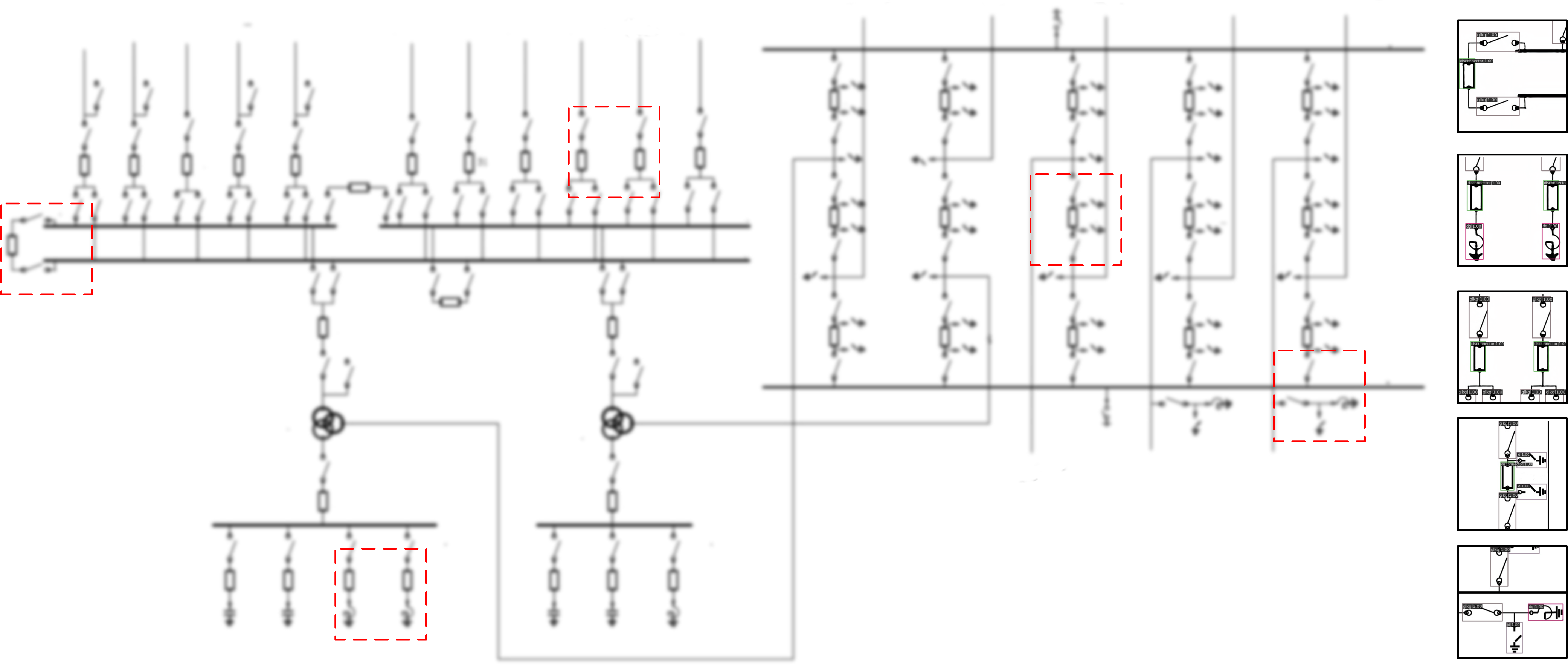}
  \caption{The restoration and recognition results of applying the proposed framework to a real-world low-quality SLD. In particular, the results of the selected windows from left to right are listed on the right-hand side in order from top to bottom.}
  \label{fig:The results of the restoration and recognition of low-quality SLD by applying the proposed framework}
\end{figure}

Moreover, a severely degraded SLD patch is used to provide an intuitive comparison between different models. In \figref{fig:Detection performance of three different methods in the face of heavily degraded drawings}, (a) shows the original high-quality SLD patch, (b) shows the result of applying the single Faster R-CNN model to the corresponding low-quality patch generated as described in Section \ref{sub:training_data_preparation}, (c) shows the result of applying independently trained models to the low-quality patch, and (d) shows the result of the proposed framework. It is evident that using a single Faster R-CNN model fails to detect any graphical symbol. In addition, collaborative training allows the two DCR symbols to be detected compared to independent training. This is likely because the independently trained ESRGAN model mistakenly restore the hollow circles on the left side of the DCR symbols as solid circles, preventing the Faster R-CNN model from correctly identifying the symbols. An intriguing thing is that both (c) and (d) appear to be well restored to the naked eyes. When measured using the structural similarity index measure (SSIM) commonly used to assess SR models, (c) even has a higher value of 0.6187 than the 0.5577 of (d). This observation emphasizes the difference between machine vision and human vision and highlights the need to use the performance of downstream tasks to evaluate the SR model. To provide a more intuitive demonstration of the proposed framework, it is finally applied to a real-world low-quality SLD, with the restoration and recognition results showcased in \figref{fig:The results of the restoration and recognition of low-quality SLD by applying the proposed framework}.

\subsection{Comparison of Different Image Restoration Models} 
\label{sub:comparison_of_different_image_restoration_models}

As discussed in Section \ref{ssub:restoration_of_ctp}, this paper implements several modifications to the original ESRGAN model. The network structure is trimmed to improve efficiency and robustness, and the activation function is changed for better performance. The gradient loss is also integrated to sharpen the restored symbol edges. Additionally, this paper proposes to use a more complex quality degradation algorithm to generate synthetic low-quality samples for training the modified ESRGAN model. The modified ESRGAN model is compared with the plain ESRGAN model, the SRGAN model, and the standard interpolation algorithm to confirm the effectiveness and superiority of the modifications. All models are trained to converge according to the method proposed in this paper. The low-quality EDs used in the experiment has random degradation levels. The average precision and recall of the eight classes of graphical symbols are listed in Table \ref{tab:Comprehensive Performance of Framework in Different Restoration Models}.

\begin{table}[!tb]
    \small
    \caption {Symbol recognition results of SLDs restored using different models}
    \label{tab:Comprehensive Performance of Framework in Different Restoration Models}
    \centering
    \begin{tabular}{ccc}
        \toprule
        Model & Precision (\%) & Recall (\%) \\ 
        \midrule
        Interpolation & 85.78 & 32.74 \\
        SRGAN & 97.27 & 85.19 \\
        ESRGAN & 97.73 & 88.93 \\
        ESRGAN + Synthetic Data & 97.94 & 98.89 \\
        Modified ESRGAN & 98.57 & 90.65 \\
        Proposed & \textbf{98.87} & \textbf{99.33} \\
        \bottomrule
    \end{tabular} 
\end{table}

The results indicate that the proposed modifications to the ESRGAN model significantly elevate both precision and recall, surpassing all the other SR models in the comparison. The increase in recall is particularly noteworthy, as it suggests that the proposed model is better equipped to handle arbitrary degradation in EDs and exhibits higher resilience. Additionally, the use of quality-degraded training data further enhances the symbol recognition performance, underscoring the robustness and effectiveness of the comprehensive approach proposed in this paper.

\subsection{Comparison of Different Recognition Models} 
\label{sub:comparison_of_different_recognition_models}

\begin{figure}[!tb]
  \centering
  \includegraphics[width=3.5in]{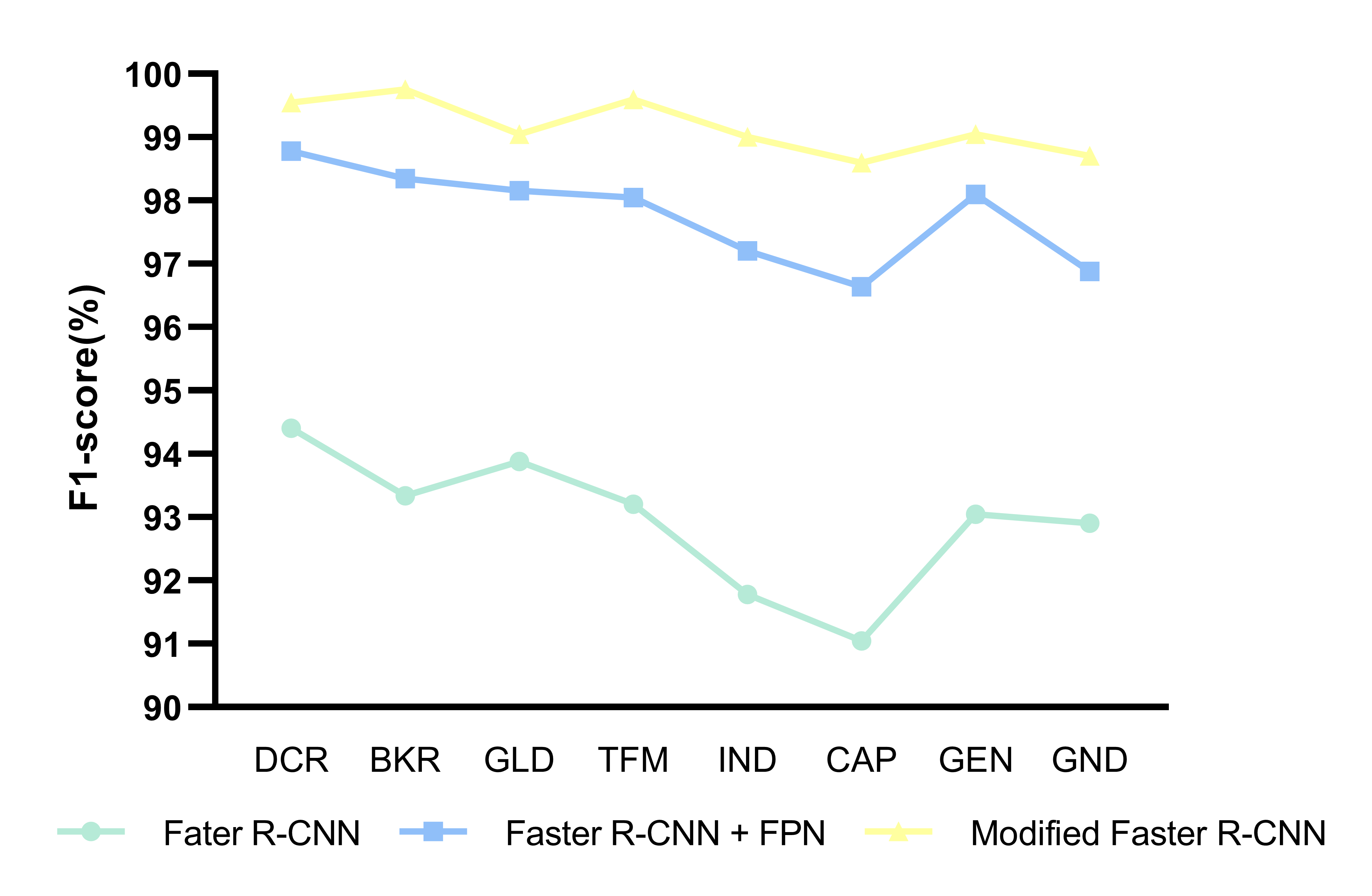}
  \caption{Recognition F1-scores for different types of graphical symbols.}
  \label{fig:Recognition results for sequential detection model improvements}
\end{figure}

As described in Section \ref{sub:graphical_symbol_recognition}, modifications are also made to the Faster R-CNN model. The Swin Transformer is used to replace the traditional CNN-based backbone, and a BiFPN architecture is implemented in the neck part. Ablation and comparison experiments are conducted to confrim the effectiveness and superiority of the proposed modified Faster R-CNN model. For comparison purposes, several object detection models commonly used in engineering practice, including the Single Shot MultiBox Detector (SSD), YOLO, and some variants of the Faster R-CNN model with different backbone structures, are trained and tested on the same data using the proposed method. The average precision and recall of the eight classes of graphical symbols are listed in Table \ref{tab:Comprehensive Performance of Framework in Different Detection Models}, with the F1-scores for each type of graphical symbol are depicted in \figref{fig:Recognition results for sequential detection model improvements}.

\begin{table}[!htb]
    \small
    \caption {Symbol recognition results of different object detection models}
    \label{tab:Comprehensive Performance of Framework in Different Detection Models}
    \centering
    \begin{tabular}{ccc}
        \toprule
        Model (backbone) & Precision (\%) & Recall (\%) \\ 
        \midrule
        YOLO (DarkNet-53) & 93.22 & 92.58 \\
        SSD (VGG-16) & 92.47 & 92.53 \\
        Faster R-CNN (ResNet-50) & 94.50 & 92.67 \\
        Faster R-CNN (ConvNeXt) & 95.03 & 93.70 \\
        Faster R-CNN (Swin Tranformer) & 95.09 & 93.96\\
        Faster R-CNN (ResNet-50) + FPN & 97.98 & 99.11 \\
        Proposed & \textbf{98.87} & \textbf{99.33} \\
        \bottomrule
    \end{tabular} 
\end{table}

The results of the comparison demonstrate that the modified Faster R-CNN model surpasses both the original Faster R-CNN model and other variants in terms of graphical symbol recognition performance. By incorporating the Swin Transformer and BiFPN, the recognition precision is enhanced by approximately 0.9\% compared to those using a CNN-based backbone and a regular FPN. Furthermore, the proposed model has much higher improvements on the F1-scores towards small symbols with similar shapes, such as CAP and GND. These findings highlight the effectiveness of the introduced modifications in enhancing the overall performance of the recognition model.

\subsection{Effectiveness Validation of Preprocessing} 
\label{sub:effectiveness_validation_of_preprocessing}

This experiment aims to validate the effectiveness of the preprocessing method in Section \ref{sub:preprocessing_of_low_quality_eds}. Ten low-quality SLDs with an average size of approximately $1400 \times 600$ are used in the experiment. The restoration and symbol recognition models remain unchanged throughout the experiment. The impact of the patch size and the involvement of patch classification on graphical symbol recognition performance and time consumption are shown in Table \ref{tab:Impact of different patch sizes and the use of patch categories on the framework}. In the table, the method ``Direct'' represents directly inputting all patches into the modified ESRGAN model for restoration, while the method ``Categorized'' represents dividing them into STPs and CTPs based on their GLCM features and restoring different types of patches separately. The time consumption is measured from the moment the first patch enters the restoration module to the completion of the last patch.

\begin{table}[!htb]
    \small
    \caption {Impact of different preprocessing methods on symbol recognition}
    \label{tab:Impact of different patch sizes and the use of patch categories on the framework}
    \centering
    \begin{tabular}{ccc}
        \toprule
        Method (patch size) & F1-score (\%) & time (s) \\
        \midrule
        Direct ($50\times50$) & 98.9 & 17.478 \\
        Categorized ($50\times50$) & 98.8 & 13.739 \\
        Direct ($100\times100$) & 99.1 & 9.733 \\
        Categorized ($100\times100$) & 99.1 & 7.592 \\
        Direct ($200\times200$) & 99.1 & 8.434 \\
        Categorized ($200\times200$) & 99.1 & 6.943 \\
        Direct ($300\times300$) & 99.1 & 8.387 \\
        Categorized ($300\times300$) & 98.7 & 7.373 \\
        \bottomrule
    \end{tabular} 
\end{table}

It can be seen that the utilization of the proposed preprocessing method offers significant advantages in terms of efficiency, as it effectively reduces time consumption while maintaining stable performance. However, it is worth noting that the patch size does not has a significant impact on the symbol recognition performance. The optimal choice cannot be simply summarized as ``the smaller the better'' or ``the larger the better'' but instead requires judicious determination based on the size of the input EDs to achieve a tradeoff between the computational cost and the overall performance.

\subsection{Practical Application of the Proposed Framework}
\label{sub:practical_application_of_the_proposed_framework}

\begin{figure}[!htb]
  \centering
  \includegraphics[width=\textwidth]{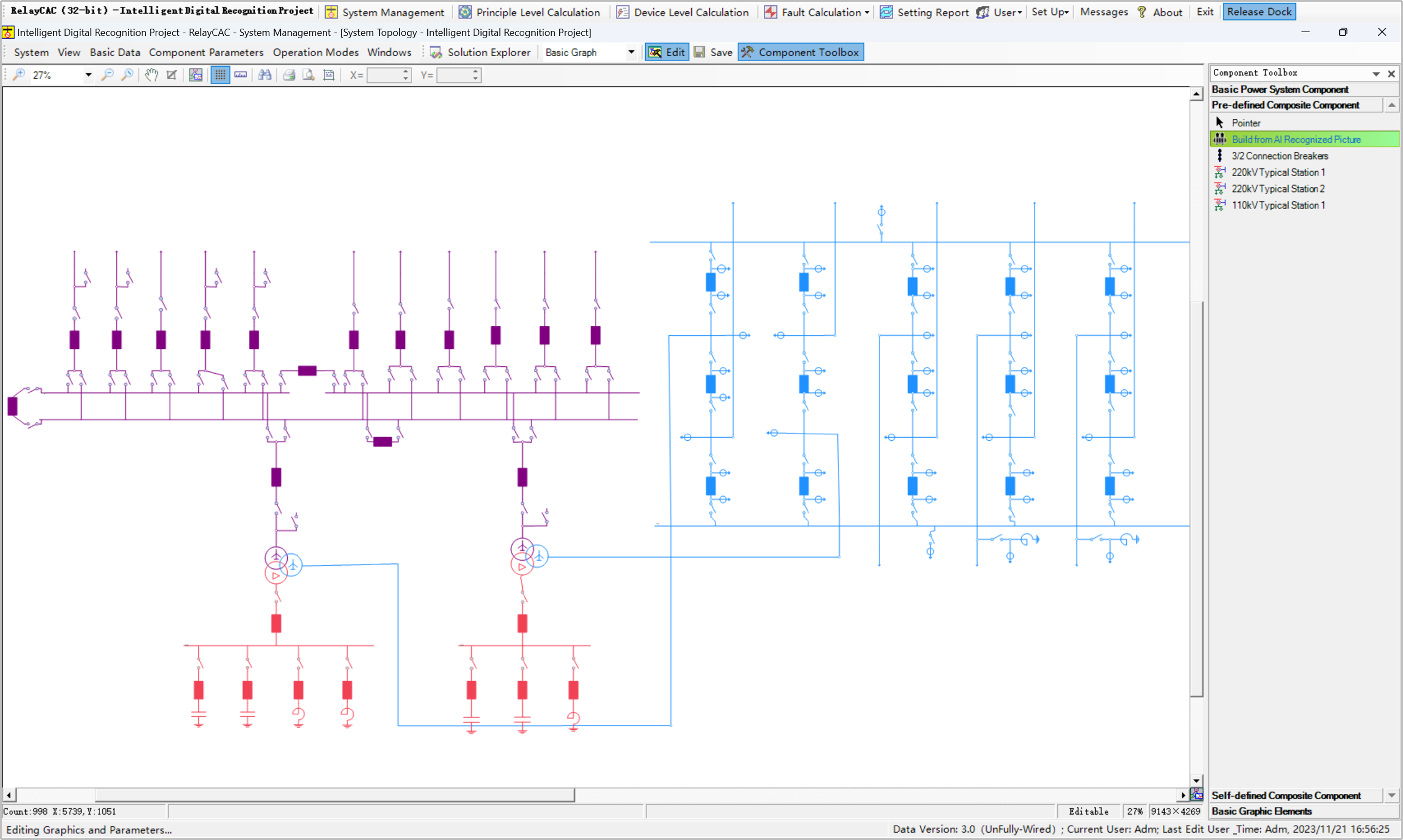}
  \caption{The substation system model automatically constructed in RelayCAC based on the digital description file generated by the proposed framework.}
  \label{fig:RelayCAC fig}
\end{figure}

The proposed low-quality ED restoration and recognition framework has been successfully integrated into RelayCAC, a commercial software utilized by over twenty provincial and regional power grid control centers in China for protective relay setting coordination. The integration is realized using XML-formatted digital description files. Specifically, the proposed framework is employed to convert the input low-quality EDs into XML files, which contain the class labels and locations of graphical symbols provided by the framework and connection lines recognized according to a previous paper by our team \cite{yang2023intelligent}. These XML description files can then be imported into RelayCAC, streamlining the system modeling process. With the proposed framework, engineers no longer need to struggle to identify blurry symbols on low-quality SLDs and manually draw them in the software. Most of the work can now be completed with a single click, enabling engineers to focus on inputting the parameters of electric equipment. The resulting substation system model, automatically constructed from an imported XML file, is illustrated in \figref{fig:RelayCAC fig}. The automatically generated system model closely mirrors the original low-quality SLD image, except that different voltage levels are represented in distinct colors based on topological analysis in the software. This application exemplifies the framework's ability to generate faithful digital representations from quality-degraded input EDs and envisions its broad application potential in other engineering fields.

\section{Conclusion} 
\label{sec:conclusion}

This paper proposes a comprehensive end-to-end framework designed to restore and recognize complex low-quality engineering drawings (EDs). The framework consists of three computer vision modules: preprocessing, image restoration, and graphic symbol recognition. GLCM features and K-means clustering are utilized to classify ED patches based on texture complexity, aiming to expedite subsequent processing steps. Simple texture patches are directly restored using computer vision operations, while complex texture patches are restored using a deep learning model modified from the Enhanced Super-Resolution Generative Adversarial Network (ESRGAN). The modifications, including reducing the network depth, changing the activation function, and introducing the gradient loss, improve the suitability for ED restoration. The restored EDs are then fed into a modified Faster Region-based Convolutional Neural Network (Faster R-CNN) model for symbol recognition, where the convolutional layers of the traditional Faster R-CNN model is replaced with the Swin Transformer and the bi-directional feature pyramid network is augmented. A multi-stage task-driven collaborative training approach is also proposed to allow EDs to be restored in a direction facilitating for the downstream recognition task. A quality-degraded training data generation method is also proposed to relieve the burden of collecting a sufficient number of low-quality EDs in practice. The proposed framework is validated through extensive experiments on real-world substation single-line diagrams. Furthermore, it is integrated into a real power engineering software application to demonstrate its practicality. Further efforts to improve the proposed framework include optimizing the structures and parameters of constituent models and exploring more effective fine-tuning strategies. It is believed that the investigation on the automatic recognition of low-quality EDs is significant for promoting digitization in various engineering fields.



\bibliographystyle{elsarticle-num} 
\bibliography{main}

\end{document}